\documentclass[journal]{IEEEtran}
\usepackage{url}
\usepackage{xcolor}
\usepackage{hyperref}

\usepackage{stfloats}
\usepackage{amsmath,amssymb}
\usepackage{mathtools}
\usepackage{xspace}
\usepackage{booktabs}
\usepackage{xcolor}
\usepackage{colortbl}
\usepackage{multirow}
\usepackage{array}
\usepackage{url}
\usepackage{makecell}
\usepackage{changepage}
\usepackage{changes}

\usepackage{framed}
\usepackage{float}
\usepackage{booktabs}
\usepackage{longtable}
\usepackage{nicefrac}
\usepackage{amssymb}
\usepackage{amsmath}
\usepackage{xspace}
\usepackage{mathtools}
\usepackage{diagbox}
\usepackage{graphics}
\usepackage{rotating}
\usepackage[algo2e,ruled,vlined,linesnumbered]{algorithm2e}

\usepackage{graphicx}
\usepackage{subfigure}
\usepackage{ifthen}
\usepackage{wrapfig}
\usepackage{balance} 
\usepackage{epsfig}
\usepackage{xcolor, colortbl}
\usepackage[normalem]{ulem}
\usepackage{dsfont}
\usepackage{hyperref}
 \hypersetup{
     colorlinks=true,
     linkcolor=blue,
     filecolor=blue,
     citecolor = blue,      
     urlcolor=blue,
     }

\begin{document}

\title{OPTION: OPTImization Algorithm Benchmarking ONtology}

\author{Ana Kostovska,
        Diederick Vermetten,
        Carola Doerr,
        Sa\v{s}o D\v{z}eroski,
        Pan\v{c}e Panov,
        and~Tome Eftimov

\thanks{Ana Kostovska (Email: ana.kostovska@ijs.si), Pan\v{c}e Panov (Email: pance.panov@ijs.si) and Sa\v{s}o D\v{z}eroski (Email: saso.dzeroski@ijs.si) are with the Department of Knowledge Technologies, Jo\v{z}ef Stefan Institute, Ljubljana, Slovenia as well as with the Jo\v{z}ef Stefan International Postgraduate School, Ljubljana, Slovenia.}
\thanks{Carola Doerr (Email: Carola.Doerr@lip6.fr) is with Sorbonne Université, CNRS, LIP6, Paris, France.}
\thanks{Diederick Vermetten (Email: d.l.vermetten@liacs.leidenuniv.nl) is with the Leiden Institute for Advanced Computer Science, Leiden, The Netherlands.}
\thanks{Tome Eftimov (Email: tome.eftimov@ijs.si) is with the Computer Systems Department, Jo\v{z}ef Stefan Institute, Ljubljana, Slovenia.}

}

\maketitle

\begin{abstract}
Many optimization algorithm benchmarking platforms allow users to share their experimental data to promote reproducible and reusable research. However, different platforms use different data models and formats, which drastically complicates the identification of relevant datasets, their interpretation, and their interoperability.
Therefore, a semantically rich, ontology-based, machine-readable data model that can be used by different platforms is highly desirable.
In this paper, we report on the development of such an ontology, which we call OPTION (\textbf{OPTI}mization algorithm benchmarking \textbf{ ON}tology). Our ontology provides the vocabulary needed for semantic annotation of the core entities involved in the benchmarking process, such as algorithms, problems, and evaluation measures.
It also provides means for automatic data integration, improved interoperability, and powerful querying capabilities, thereby increasing the value of the benchmarking data.
We demonstrate the utility of OPTION, by annotating and querying a corpus of benchmark performance data from the BBOB collection of the COCO framework and from the Yet Another Black-Box Optimization Benchmark (YABBOB) family of the Nevergrad environment. In addition, we integrate features of the BBOB functional performance landscape into the OPTION knowledge base using publicly available datasets with exploratory landscape analysis. Finally, we integrate the OPTION knowledge base into the IOHprofiler environment and provide users with the ability to perform meta-analysis of performance data.
\end{abstract}


\IEEEpeerreviewmaketitle

\section{Introduction}

Evolutionary computation and black-box optimization in general are fast-growing fields that have made tremendous progress recently. Due to the numerous applications in engineering, artificial intelligence, and beyond, new optimization algorithms are constantly being developed, making it impossible for researchers and practitioners in the field to keep up with all the new developments. On the other hand, data sharing has gained significant acceptance in recent years. Nowadays, it is de facto standard to publish research results and data on publicly accessible data repositories, whenever possible, to promote their reusability. However, while data sharing undoubtedly helps to achieve this aim, there are unfortunately no common standards for \emph{which} data to record, nor \emph{how} to store it. Consequently, the storage, sharing, and reusability of benchmark optimization data is challenging because there are different data formats that are only partially compatible. In the context of numerical optimization, for example, there are several important benchmarking tools, each with its own way of storing performance data. While each of these individual data formats is internally consistent, they handle the details of data storage differently. Interoperability of data from different sources is therefore limited unless explicit conversions are made. Management of benchmark data is even more challenging if we consider the “Reproducibility guidelines for AI research”~\cite{gundersen2018state,carter2019pragmatic}, where various ACM reproducibility concepts are presented: i) repeatability (same team, same experimental design), ii) reproducibility (different team, same experimental design), and iii) replicability (different team, different experimental design). These guidelines are also discussed in the context of evolutionary computation ~\cite{ManuelBP21reproducibilityTELO}.

In addition to performance data, problem landscape data also play an essential role in benchmark studies. Problem landscape data represent the characteristics of an optimization problem instance on which an optimization algorithm is run to obtain the performance data. One way to represent the characteristics of problem instances is to use exploratory landscape analysis (ELA)~\cite{mersmann_exploratory_2011}. ELA provides low-level features computed from a sample of observations for a given problem instance. It has already been shown that representing the characteristics of problem instances using ELA features can provide promising results in automated prediction of algorithm performance~\cite{eftimov2021personalizing,jankovic2021towards}, automated algorithm selection~\cite{kerschke_automated_2019,MunozSurvey15}, and automated algorithm configuration~\cite{BelkhirDSS17}. However, calculating the ELA features can be a computationally intensive step. To avoid recomputation 
the ELA features should be made reusable and interoperable. 

In summary, while we see an increasing amount of data that could be used to select or even design algorithms in an informed way, we also see increasing complexity in identifying and processing relevant data, leading to biased comparisons and reduced transferability of knowledge. 
A well-established solution to the problem of managing such complex collections of data is the use of ontologies. An \emph{ontology} is a formal, explicit specification of a shared conceptualization \cite{smith2012ontology}. The explicit semantic assumptions used in representing domain knowledge ensure a shared understanding of the domain, while the use of formal knowledge representation mechanisms makes it machine-readable. Thus, an ontology can be viewed as formal representation of domain knowledge, while the databases semantically annotated with ontology terms are known as \emph{knowledge bases (KB)}. The Gene Ontology (GO)~\cite{gene2019gene} is a major success story for the development and use of ontologies for data linkage and integration. Due to the large pool of data annotated with GO, the expressive power of the ontology and its automated reasoning services, new biological knowledge has been derived (e.g., classifying protein phosphatases \cite{wolstencroft2005using}).

Apart from structuring domain knowledge in a principled way, ontologies improve data sharing and reusability by providing mechanisms for explicitly specifying provenance information of different resources \cite{hartig2009provenance}. Provenance information is the type of information that describes the origin of a resource, such as who created the resource, when it was published, and what license applies to its use.

We believe that efforts to formalize knowledge about benchmark performance and problem landscape into an ontology would be very worthwhile and would help the community address the challenges of data integration, knowledge sharing, and data reusability. To this end, we have developed the OPTimization Algorithm Benchmarking ONtology (OPTION) and demonstrated the utility of ontology-based systems for data management.

\noindent\textbf{Our contribution:} 
The main contribution of this paper is the OPTION ontology. The ontology we propose was designed to standardize and formalize knowledge from the domain of benchmarking optimization algorithms, where an emphasis is put on the representation of data from the benchmark performance and problem landscape space. 

The OPTION ontology addresses several key challenges in data representation: (1) The different data formats in which benchmarking data is stored; (2) Different implementations of algorithms and problems under the same name; (3) Same algorithms or problems, but with different names; and (4) The complexity of the benchmark data.

There are three other technical contributions of the paper in support of the main contribution that constitute the ontology-based system for data integration: (i) the OPTION KB -- semantic annotations of benchmark performance data from the BBOB~\cite{COCOdata} and Nevergrad~\cite{nevergrad} platforms of over 200 different optimization algorithms and semantic annotations of publicly available collections of problem landscape data;
(ii) an endpoint for querying the OPTION KB via a graphical user interface implemented in the IOHprofiler environment~\cite{IOHprofiler,IOHanalyzer}; and 
(iii) a curator-assisted pipeline that enables 
to extension the OPTION ontology and OPTION KB. \\
 
 \noindent\textbf{Organization:} This paper is organized as follows. First, in Section~\ref{sec:background}, we describe what ontologies are and what technologies are used for their processing. We also discuss currently available ontologies for optimization and evolutionary computing. Next, in Section~\ref{sec:data_formats}, we focus on the challenges of integrating performance and problem landscape data and discuss how these challenges can be addressed with ontologies. In Section~\ref{sec:ontology}, we describe our core contribution, the OPTION ontology. Furthermore, in Section~\ref{sec:use-cases}, we describe three data sources (BBOB, Nevergrad and ELA) as use-cases for semantic annotation using the OPTION vocabulary and constructing the OPTION KB, while in Section~\ref{sec:system}, we describe the OPTION system for semantic data management and integration. Finally, we conclude the paper with a summary of contributions and plans for future work in Section~\ref{sec:conclusions}.

\section{Background \& Related work}\label{sec:background}
\subsection{Ontologies as representational artefacts}
In philosophy, the term ontology refers to the study of being. However, in the context of computer science, \textbf{ontologies} are ``explicit formal specifications of the concepts and relations among them that can exist in a given domain''~\cite{gruberOntology}. Ontologies provide the basis for a unambiguous, formal representation of domain knowledge usually approved by experts in the domain. 
They take an object-oriented view of modeling and include notions like: (1) \emph{classes,} a set of semantically defined concepts; (2) \emph{individuals,} instances of classes; and (3) \emph{properties,} binary relations used to associate the classes and/or the individuals. 

It is important to distinguish between two types of ontology statements: TBox and ABox. \emph{TBox} statements form the ``terminology component'' and describe the domain of interest by defining the classes and properties that form the vocabulary of the ontology (analogous to object-oriented classes). \emph{ABox} statements are the ``assertion component" facts associated with the TBox (analogous to instances of object-oriented classes). By assigning semantic meaning to knowledge facts and explicitly linking them to ontology terms, we perform the task of semantic annotation. 
ABox statements form is a set of semantic annotations or a \emph{knowledge base (KB)}. All KBs corresponding to an ontology are interoperable, which means that distributed, heterogeneous systems and databases can easily interconnect and exchange information. Using the same ontology when annotating data and creating KBs facilitates automatic data integration, where classes and properties defined in the ontology serve as connecting point of the KBs.

Ontologies provide the means for knowledge and data representations that are semantically understandable and available in machine-processable form. Thus, ontologies play a crucial role in sharing a common understanding of information structure among people or software agents. Numerous applications from different domains that involve big data handling are based on ontology as a data model that further allows domain knowledge analysis. Such applications can be found in biomedicine~\cite{gene2001creating}, food and nutrition~\cite{dooley2018foodon}, environmental studies~\cite{buttigieg2013environment}, etc. In recent years, ontologies are also used to represent computer science domains, such as the domains of data mining (DM) and machine learning (ML)~\cite{panov2014ontology}.

\subsection{Semantic web technologies} 
Ontologies as informational artifacts have long been used to develop machine-readable, semantically interoperable data, which is essentially the central goal of the Semantic Web~\cite{semanticweb}. 

The Semantic Web has been trying to achieve this goal via semantic annotation of data found on the web with terms defined in ontologies, intended to give data a well-defined meaning. As a result, many technologies have been developed, including RDF, RDFS, triplestores, SPARQL, etc. Here, we describe the Semantic Web technologies relevant to developing the OPTION ontology and knowledge base.

RDF\footnote{Resource Description Framework:\url{https://www.w3.org/RDF/}} is a standard data/metadata exchange format. It provides a simple data/metadata format for expressing statements using RDF triples. RDF triples are composed of a subject, a predicate, and an object. Each triple represents one fact about a resource (subject) and a property (predicate) with a given value (object), e.g., "John knows Peter.", where, John is the subject, Peter is the object, and 'knows' is the link/predicate.

RDF triples are stored in \emph{triplestores,} a specific type of NoSQL database, similar to graph databases (databases that use graph structures to represent data). Triplestores can store trillions of RDF records, which makes them applicable in the Semantic Web context. In addition, triplestores offer several advantages over relational databases, including flexibility (data can be easily altered since there is no predefined data schema), easy import/export of triples, efficient querying, and easy sharing. There are many implementations of triplestores, and one of the most used ones is Apache Jena TDB\footnote{Apache Jena TDB: \url{https://jena.apache.org/documentation/tdb/}}, which is a part of a larger open-source framework for Semantic Web applications. SPARQL\footnote{W3C SPARQL 1.1: \url{https://www.w3.org/TR/sparql11-query/}} is an RDF query language able to retrieve and manipulate data stored in RDF format.

RDF Schema (RDFS)\footnote{W3C RDF Schema 1.1: \url{https://www.w3.org/TR/rdf-schema/}} is another semantic technology standard that is an extension of the RDF data model and provides essential elements for describing ontologies, such as classes and properties (relations). The Web Ontology Language (OWL)\footnote{W3C OWL2: \url{https://www.w3.org/TR/owl2-overview/}} is a collection of representation languages for authoring ontologies with different levels of expressivity. Data stored in a graph format (e.g., as an RDF graph) integrated into an ontology is commonly referred to as a \emph{knowledge base.}

Ontologies as computational artifacts are usually based on Description Logic (DL) as a knowledge representation formalism~\cite{baader2003description}. This logical component allows knowledge to be shared meaningfully at both machine and human levels. Also, an immediate consequence of having formal ontologies based on Description Logic is that they can be used in a variety of reasoning tasks and inference of new knowledge. For example, several reasoning engines can infer new knowledge from OWL ontologies, such as Hermit~\cite{hermit} and Fact++~\cite{tsarkov2006fact}.

\subsection{Ontologies for optimization and evolutionary computing} 
Several efforts have been made at conceptualizing different aspects of domain knowledge about evolutionary computation. The Evolutionary Computation Ontology (ECO) has been developed to model the relations between algorithm settings (i.e., solution encoding, operators, selection, and fitness evaluation) and different types of problems~\cite{yaman2017presenting}. It is focused on describing the properties of algorithms, which can be especially helpful for teaching EA-related topics. In the domain of multi-objective optimization, the Diversity-Oriented Optimization Ontology has been developed, including a taxonomy of algorithms concerning the diversity concept in different search operators~\cite{basto2017survey}. Complementary to the diversity concept, the Preference-based Multi-Objective Ontology (PMOEAs) has also been proposed to model the knowledge about preference-based multi-objective evolutionary algorithms~\cite{li2017building}. 

The above ontologies have a strong focus on specifics, resulting in classifications of algorithms that allow users to ask only about high-level relations. For example, finding algorithms that use a specific type of operator, finding algorithms that can solve problems from a particular class, and finding algorithms that have been applied to a specific engineering problem. What is missing are ontologies that add semantics to available benchmark data, so that high-level relations and conclusions can be drawn from the ontologies. 

\section{Towards Integrating Performance \& Problem Landscape Data }\label{sec:data_formats}

\subsection{Performance Data}
There exist many different benchmarking platforms for optimization, each with their own way to store performance and algorithm data. Three main approaches to the storage of performance data are described below:
\begin{itemize}
    \item \textbf{Csv-based}: The data is stored as a single file per experiment in a csv-based format, where each column represents a performance measure or other meta-information. An example is the format used in Nevergrad~\cite{nevergrad}. This allows for storing data on many different functions/problems into a single file, with the drawback that the granularity of the data is often limited.
    \item \textbf{Textfile-based}: The data is separated into a single file per function/problem, where the meta-information is delimited in some way, followed by the performance information. This format is easily extendable and human-readable, but it can be hard to work with when files become large. An example of this format is used by the SOS platform~\cite{SOS}.
    \item \textbf{COCO/IOH-like}: 

    The data is separated into multiple files and folders: generally, folder structure splits along algorithms and functions/problems. Each folder then contains a file with meta-information about the runs, with links to the files where the raw performance data is stored. This structure makes it easy to find the data sought, but the different links to the files can be an obstacle for practitioners who are not used to this format. Variants of this data format are used by COCO~\cite{hansen2020coco} and IOHprofiler~\cite{IOHprofiler}.
\end{itemize}
As mentioned, these data formats has its advantages and disadvantages. While there are some commonalities between different methods, the particularities in handling meta-data make interoperability of the data from different sources challenging. Furthermore, these differences lead to more limited post-processing functionalities available to the users of these platforms since they are only compatible with those tools that support their particular data format. While these tools are slowly becoming more interoperable,

this process requires significant effort from the developers of the individual tools to make sure all data formats are fully supported.
A common data structure would be useful to the benchmarking community to avoid each developer having to do this individually. 

In order to create such a common structure, it is crucial to identify the core components of performance data that would be needed in the analysis. We will specify these components in Section~\ref{sec:ontology}. Then, an explicit conversion needs to be made for each data format, which extracts these properties from the original data format. Some previous efforts show that it is possible to make such conversions and then jointly analyze data which has been originally stored in different data formats. In particular, performance data from Nevergrad, COCO, IOHexperimenter, and from the SOS platform can be conveniently analyzed through the IOHanalyzer~\cite{IOHanalyzer}.

\subsection{Problem Landscape Data}
The problem space includes optimization problems that can belong to different problem classes~\cite{TBB20benchmarking}. One problem class can have many different problem instances. 
For numerical optimization, for example, the differences can result from transformation processes in the search space (such as translation, shifting, and/or scaling of the problem instance). Transformations in the objective space are also common.  

Landscape analysis aims at characterizing problem instances by providing features obtained by applying mathematical and statistical methods. For example, exploratory landscape analysis (ELA)~\cite{mersmann_exploratory_2011}, was designed to support the design of black-box optimization algorithms through a set of recommendations based on Machine Learning (ML) to obtain algorithms that best fit the problem at hand. 

The main objective of ELA is to capture the characteristics of optimization problems with a set of features, referred to as ELA features. These features are then used as input to ML pipelines that produce recommendations that guide algorithm design. As we are dealing with black-box optimization, these features must be calculated from an  (ideally small) set of samples of the problem instance. The ELA features can be computed using the R-package \emph{flacco}~\cite{flacco2019}, which contains 343 ELA features grouped into 17 feature sets~\cite{kerschke_automated_2019}. In this study, we use the dataset ~\cite{quentin_renau_2020_3886816} containing already calculated ELA features for some COCO benchmark problems.

\subsection{Domain challenges for data integration} 
A source of complexity in recording performance data from black-box optimization is that we typically do not use a single performance measure. Instead, we are interested in analyzing algorithm performance from different perspectives: small vs. large budgets, the time needed to identify solutions that meet specific quality criteria, the robustness of the algorithm in search and performance space, etc.~\cite{TBB20benchmarking}. 

To enable such detailed analyses, researchers often record performance data in a multi-dimensional fashion, spanning at least the time elapsed (measured in terms of CPU time and/or function evaluations), solution quality, and robustness. We may also be interested in how dynamic parameters evolve during the optimization process, in which case we record their values along with the performance data. Both requirements add another level of complexity to the data formats and may explain why they differ so much in practice.

There are several other factors that further complicate the interoperability and re-usability of publicly available performance data from different benchmarking experiments:
\begin{itemize}
    \item Most black-box optimization algorithms are, in fact, families of various algorithm instances. They can be selected by specifying the (hyper-)parameters of the algorithm and/or the operators (e.g., one may speak of Bayesian optimization regardless of the internal optimization algorithm that is used to search the surrogate model, or one may use different acquisition functions, different techniques to build the surrogate, etc.). 
    Different configurations can lead to drastically different search behavior (and hence performance), and it is crucial to associate the recorded data to the appropriate algorithm instance and not only to an algorithm family. However, this is not an easy task, as it can happen that essentially the same algorithm is published under different names~\cite{StuetzleGreyWolf, Sorensen15} for recent examples and a discussion, respectively).  
    \item A similar issue appears on the problem side. Different instances of the same problem can be of different complexity, and it is not always clear which problem instances were used within a given benchmark study. In addition, some benchmarking suites automatically rotate, shift, permute, or translate the problem instances, to test specific unbiasedness characteristics and the generalizability of the algorithms. Other suites do not do this (e.g., because the variable order or absolute values carry some meaning) but still refer to problems of different complexity under the same name. As for the algorithms, we can also have the same problem appear under different, possibly multiple, names. The \textsc{OneMax} problem, for example, is sometimes called \textsc{CountingOnes}, \textsc{OnesMax}, the Hamming distance problem, or Mastermind with 2 colors. All these names refer to the same problem. 
\end{itemize}

Identifying such issues cannot (as yet) be done automatically but require human expertise to annotate the data correctly. 

While this requires a significant amount of effort for the large amounts of currently available benchmarking data, we aim for the procedure to convert from different data formats to be automated where possible (e.g., by involving the authors of the different benchmarking platforms) and clearly structured where not. In the future, this would then become second nature when introducing a new algorithm / problem / experimental setup, allowing the data ontology to grow organically. The creation of reproducible and readily available data will eventually benefit the optimization community as a whole, so the efforts invested to achieve this goal would be very much worthwhile.

\subsection{Addressing data integration challenges with ontologies}
\label{sec:competency}
In order to develop ontology-based solutions for integration of benchmark performance and problem landscape data from different data sources, we need an ontology that formalizes the knowledge in the domain of interest. 

The ontology should cover the competency questions presented in Table~\ref{tab:competency}.\begin{table}[t]
    \caption{Ontology competency questions}
    \label{tab:competency}
    \centering
    \begin{tabular}{m{0.1cm}m{7.5cm}}
    \hline
         $N^0$ &  Competency question\\
    \hline
        1 & Which problem instances belong to a given benchmark problem?\\ 
    2 & What is the provenance data related to a given benchmark study? \\
    3 & Which algorithms are benchmarked in a given study?\\
    4 & What are the values of ELA features of a given problem instance calculated on a sample obtained by using a given sampling technique?\\ 
    5 & What was the fitness achieved for a given benchmark problem after a fixed amount of function evaluations?\\ 
    6 & How many function evaluations were needed to reach a given fitness target?\\ 
    7 & Which algorithm(s) achieve the best performance given a fixed number of function evaluations?\\
    \hline
    \end{tabular}

\end{table}

Once the ontology is defined, it can be used by different ontology-based systems and/or benchmark platforms as a common vocabulary for semantic annotation of the data. Benchmark platforms can keep their proprietary data format. As long as they annotate the data with semantic metadata and store the annotations in a semantic data store compliant with the proposed ontology, the data would be automatically interoperable with other knowledge bases and platforms that follow the same protocol for data management. 

The main goal of this work is to design an ontology for semantic annotation of benchmark performance and problem landscape data as well as to design a prototype data management system that enables data integration and reusability with the use of the ontology as a common vocabulary.  

This goal is operationalized as a set of requirements that an ontology-based system for data integration should fulfill. The requirements are presented in Table~\ref{tab:requirements}.
\begin{table}[h]
\caption{Requirements of the ontology-based system.}
    \label{tab:requirements}
    \centering
    \begin{tabular}{m{0.1cm}m{7.5cm}}
    \hline
         $N^0$ &  Requirement\\
    \hline
     1 & Semantically annotate benchmark performance and problem landscape data from different benchmark platforms and different problem test suites with ontology defined terms.\\
    2 & Store semantic annotations in a specialized semantic data store.\\
    3 & Load and query benchmark performance data from experiments performed using the same or a different system/platform.\\ 
    4 & Load and query problem landscape data for benchmark problems defined in the same or different test suite.\\ 
    5 & Load and query provenance information associated with the benchmark studies.\\ 
    6 & Allow members of the community to extend the ontological conceptual model to cover parts of the domain knowledge missing in the latest active version of the ontology. \\
    7 & Allow members of the community to upload their performance data and problem landscape data to extend the system's knowledge base.\\
    \hline
    \end{tabular}
\end{table}

\section{The OPTION ontology}\label{sec:ontology}
We developed the OPTION ontology with the primary goal of formalizing knowledge about benchmarking of optimization algorithms, emphasizing formal representation of data from the performance and problem landscape space. Thus, OPTION offers a comprehensive description of the domain covering the benchmarking process and the core entities involved in the process, such as optimization algorithms, benchmark problems, and evaluation measures. 

\begin{sloppypar}
\subsection{Ontology design}
The design of the ontology was governed by the competency questions listed in Table~\ref{tab:competency}. In the ontology design phase, we followed best practices of ontology engineering, i.e., the OBO Foundry principles~\cite{smith2007obo}, which ensure interoperability with other external ontologies that follow the same design principles. Our proposed ontology adheres to the single inheritance principle, does not contain any orphan classes, and heavily reuses formally defined relations from the Relations Ontology (RO)~\cite{smith2005relations}. Furthermore, we aligned OPTION with BFO~\cite{arp2015building}, a widely-used upper-level ontology, which served as a template to organize the class hierarchy. The ontology is also aligned with mid-level ontologies, such as IAO (the Information Artifact Ontology)\footnote{IAO ontology: \url{https://github.com/information-artifact-ontology/IAO/}} and OBI (Ontology of Biomedical Investigations) \cite{bandrowski2016ontology}. Finally, we reused classes from external ontologies, such as OntoDT (Generic ontology of datatypes)~\cite{panov2016generic}, and SIO (Semanticscience Integrated Ontology)~\cite{dumontier2014semanticscience}.
\end{sloppypar}

\subsection{Implementation and availability} 
OPTION has 347 classes, 3749 axioms, and 479 subClassOf axioms. We used the \texttt{rdfs:label} annotation property to provide human-readable English labels. The ontology is implemented as an OWL 2 DL ontology with SROIQ(D) level of expressiveness. For the development, we used Protégé~\cite{noy2003protege}, an open-source ontology-development and knowledge-acquisition environment. The ontology is publicly available at a GIT repository\footnote{OPTION at GIT: \url{https://github.com/KostovskaAna/OPTION-Ontology}}and on BioPortal\footnote{OPTION at BioPortal: \url{https://bioportal.bioontology.org/ontologies/OPTION-ONTOLOGY}} \cite{noy2009bioportal}, the largest public repository of ontologies. 

\subsection{Ontology layers}

\begin{figure}[t]
    \centering
    \includegraphics[width=\linewidth]{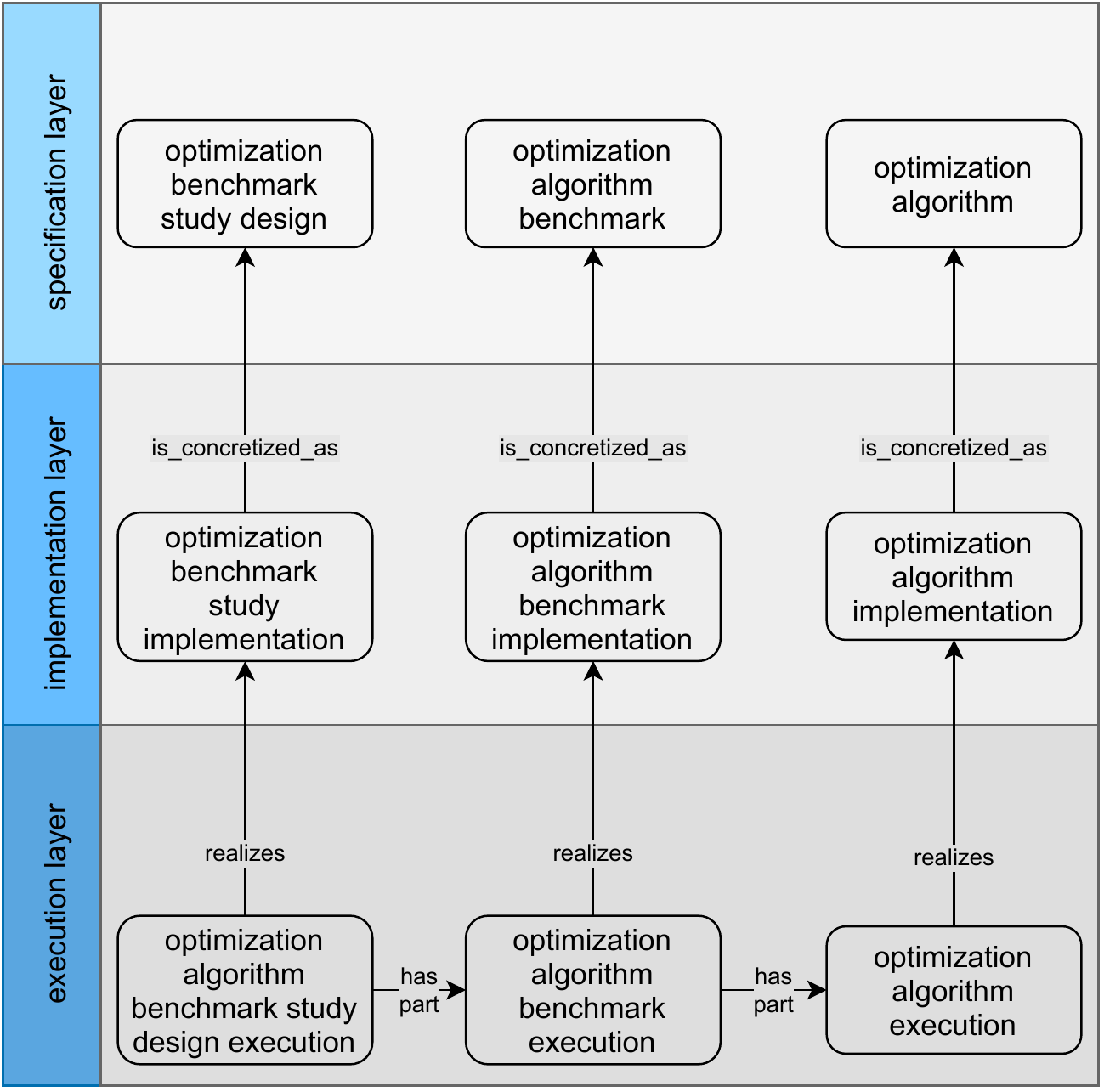}
    \caption{The specification-implementation-execution design pattern as used in the OPTION ontology.}
    \label{fig:pattern}
\end{figure}

As mentioned above, the domain classes from the OPTION ontology are aligned with middle- (IAO and OBI) and upper-level (BFO) ontologies. Orthogonally to that, at each level (domain, middle, and upper), where appropriate, we implement the specification-implementation-execution ontology design  pattern~\cite{lawrynowicz2017algorithm}. This pattern helps us describe the different aspects of one concept: For an optimization algorithm, we define three conceptually different classes, i.e., \textit{optimization algorithm}; \textit{optimization algorithm implementation}; and \textit{optimization algorithm execution} to represent general information about the algorithm; characteristics of its implementation; and information about the process of its execution, respectively (see Figure~\ref{fig:pattern}). Similarly, the \textit{optimization algorithm benchmark study design execution} and \textit{optimization algorithm execution} classes are modelled using the same design pattern (see Figure~\ref{fig:pattern}). Such representation allows us perspectivism and relativism, as in some cases, we are only interested in a specific view (e.g., implementation) of the modeled concept, while the rest can be irrelevant.  
\begin{figure*}[t]
    \centering
    \includegraphics[width=\textwidth]{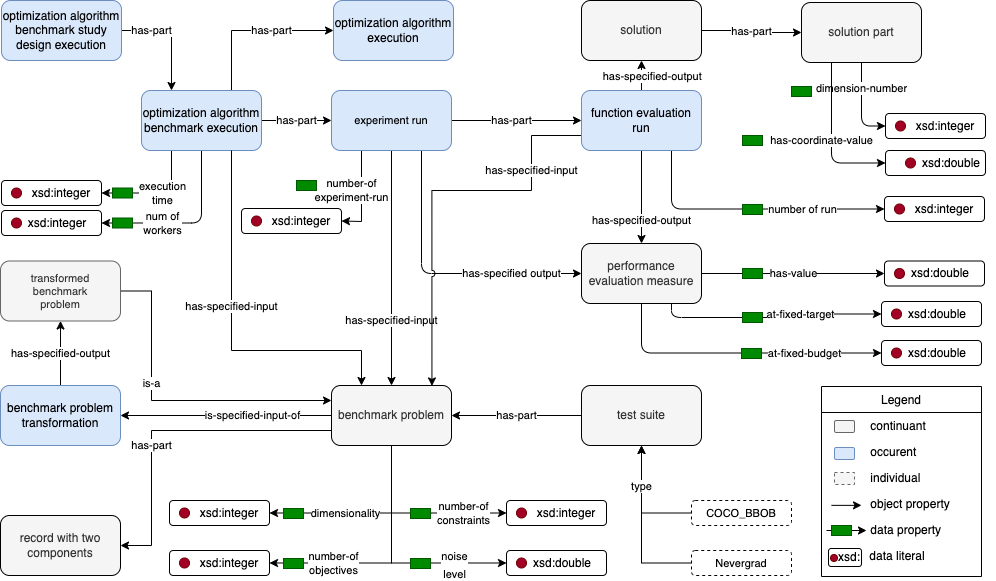}
    \caption{The core entities in the OPTION ontology and their relations.}
    \label{fig:core}
\end{figure*}

\subsection{Core entities}
The OPTION ontology is structured around several objects (i.e., continuants): \textit{benchmark problem}\footnote{In the remainder of this paper, we will refer to the ontology classes in OPTION in \textit{italic} font, while the relations between the classes will be written in \texttt{typewriter} font. }, \textit{optimization algorithm}, \textit{function evaluation}, \textit{solution}, \textit{performance evaluation measure} and others, as well as processes (i.e., occurents) in which these entities participate such as \textit{optimization algorithm benchmark study design}, \textit{optimization algorithm benchmark execution}, \textit{experiment run}, and \textit{function evaluation run}. The notion of continuants and occurents come from the BFO top-level ontology. More specifically, BFO divides all classes/entities into those two disjoint categories. Subclasses of the \textit{continuant} class are objects (including information artifacts), while subclasses of the \textit{occurent} class are processes as they can be extended through time. 

We will briefly describe how optimization entities are semantically defined within the OPTION ontology. Note that the ontology model is not presented in complete detail due to space limitations. 
For visual exploration of the ontology classes, we refer the reader to: \url{https://service.tib.eu/webvowl/#iri=https://raw.githubusercontent.com/KostovskaAna/OPTION-Ontology/main/OntoOpt.owl}. However, to explore the ontology fully (not just classes but also axioms), we advice the reader to load the OPTION ontology (the .owl file) in Protégé~\cite{musen2015protege}. 
The structure of the OPTION core entities in the ontology and their relations are presented in Figure~\ref{fig:core}.

\begin{sloppypar}
The \textit{benchmark problem} class is part of a test suite and it can undergo a process of \textit{benchmark problem transformation} (e.g., shift and scale). The transformed benchmark problem inherits all the properties of a benchmark problem. Thus, it is represented as its subclass in the ontology via the \texttt{is-a} property. For each benchmark problem at instance level we can define data properties such as \texttt{dimensionality}, \texttt{number-of-objectives}, \texttt{number-of-constraints}, and \texttt{noise level}. For the representation of datatypes on the decision and objective space, we imported the \textit{record with two components} class from OntoDT. The first component is associated with datatypes on the decision space, the second with datatypes on the objective space. Since different test suites can have different benchmark problems or variations of the same, we use the \textit{benchmark problem} class as a root class to build the taxonomy of benchmark problems for each test suite separately. For example, 24 benchmark problems constitute the taxonomy of BBOB benchmark problems.
\end{sloppypar}

To represent the study design and study design execution concepts in the context of benchmarking optimization algorithms, we defined the \textit{optimization algorithm benchmark study design} and \textit{optimization algorithm benchmark study design execution} classes as specializations of classes already defined in the OBI ontology. 
Since keeping track of the provenance information related to each study is a very important aspect in the context of reproducibility and traceability of experiments, we imported a number of properties from the well known Dublin Core~\cite{dublin2012dublin} vocabulary and metadata schema, such as \texttt{dc:identifier}, \texttt{dc:title}, \texttt{dc:date}, and \texttt{dc:creator}, to name a few.

In one study, we can benchmark a set of optimization algorithms. This relation is captured with the \texttt{has-part} transitive object property between the \textit{optimization algorithm benchmark study design execution} and \textit{optimization algorithm benchmark execution}, which represents execution of each individual algorithm (see Figure~\ref{fig:core}). 

The \textit{optimization algorithm benchmark execution} class includes specification of the input(s) (i.e., the benchmark problem) and its sub-processes. The execution process is composed of two sub-processes: \textit{optimization algorithm execution} and \textit{experiment run}. Here, we also specify details about the execution, such as the execution time and the number of workers (when parallelization is allowed).

The definition of the {experiment run} class includes specifics about the input(s) of the process (i.e., the benchmark problems) and output(s) (i.e., performance evaluation measure) of the execution process. In the ontology, various performance evaluation measures have been represented. These include the measured fitness, the best-measured fitness, and the noise-free fitness measure. In addition, the ontology also supports the representation of performance data at the level of function evaluations. For that purpose, we use the same \textit{performance evaluation measure} as presented above. Note that we associate information about the function evaluation runs where applicable, as not all benchmark data is given at this level of granularity. Finally, performance can be measured in a fixed-target or fixed-budget scenario. 

Moreover, we also define a \textit{solution} as an output of each \textit{function evaluation run} process. Each solution is broken down into multiple parts, and for each solution part, we represent its location via the \texttt{has-coordinate-value} data property. The number of solution parts depends on the dimensionality of the problem. 

\begin{figure}[h]
    \centering
    \includegraphics[width=\linewidth]{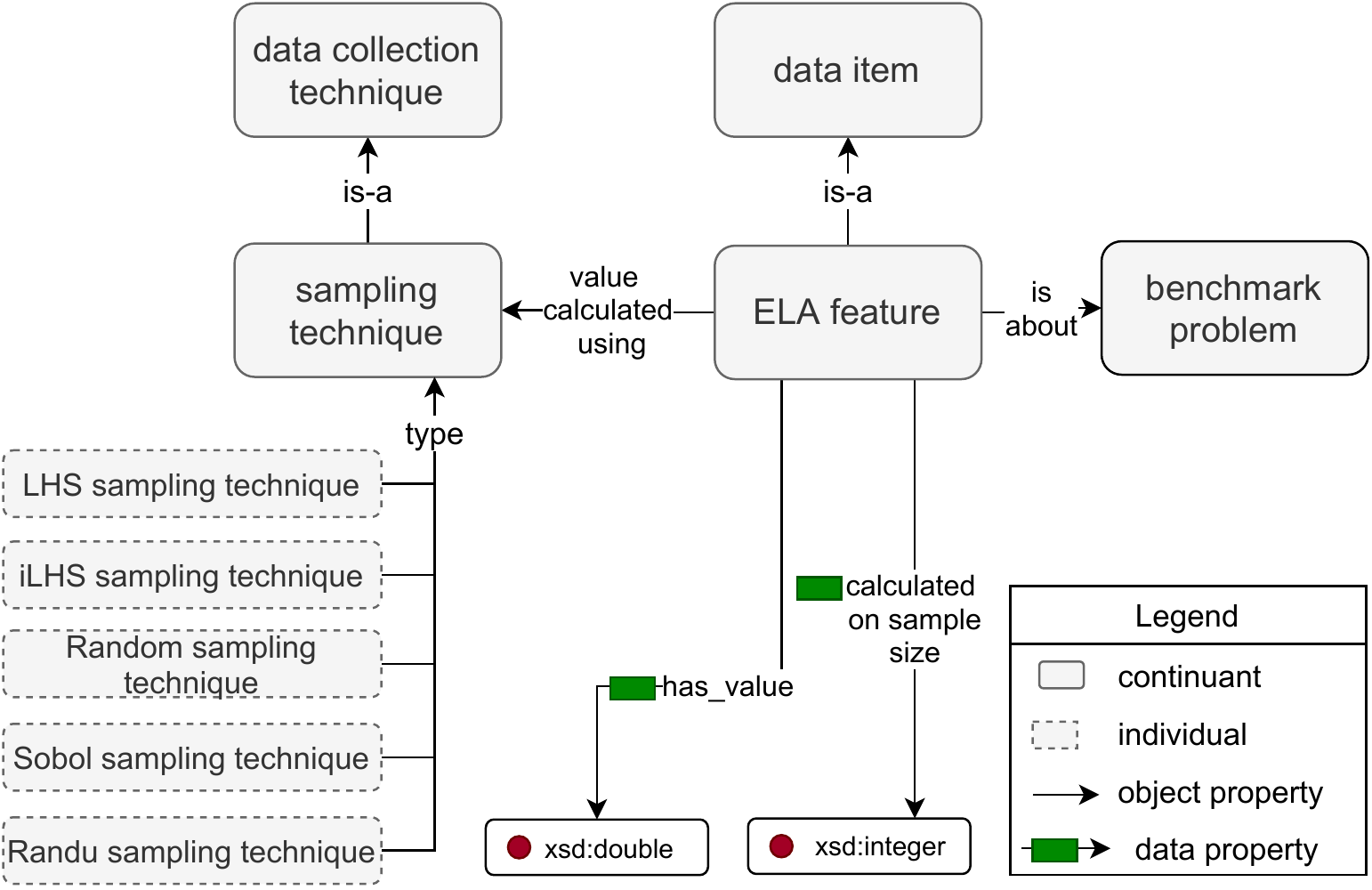}
    \caption{Representation of ELA features in the OPTION ontology.}
    \label{fig:ELA}
\end{figure}

Finally, the problem landscape space is represented with ELA features. \textit{ELA features} in the OPTION ontology are defined as \textit{data items} (see Figure~\ref{fig:ELA}). They are linked with the corresponding \textit{benchmark problem} via the \texttt{is-about} relation. The \textit{benchmark problem} class in Figure~\ref{fig:core} is the same one as in Figure~\ref{fig:ELA} and it connects the two figures. Since the ELA feature value depends on the \textit{sampling technique} and the \textit{sample size}, this information is also included in the ontology. We have already included five sampling techniques that are most common in the literature. However, the list can be extended with other sampling techniques on-demand.

\section{Use cases}
\label{sec:use-cases}
To demonstrate the benefits of using a common ontology for semantic annotation of data and to address some of the domain challenges for data integration, we consider 3 different use cases or data sources. The BBOB and Nevergrad benchmark suites are out use cases for annotation of performance data, and a large set of publicly available ELA data a use case for annotation of problem landscape data. We use the OPTION ontology as a formal vocabulary for semantic annotation of performance data and problem landscape data.

\subsection{BBOB} 
Since 2009, annual workshops have been organized around the benchmarking of derivative-free black-box optimization algorithms with the COCO environment~\cite{hansen2020coco}. We consider the BBOB single-objective benchmark suite~\cite{bbob2009}, which   
consists of 24 single-objective benchmark functions. 
Some of the data generated during these workshops is freely available~\cite{COCOdata}. It covers results for the problems of different dimension $D\in\{2,3,5,10,20,40\}$.

We use the BBOB data that includes the algorithms from the 2009-2020 workshops in this use case. The data includes 226 algorithms, which we semantically annotate using the OPTION ontology. For each of the 226 algorithms, semantic annotations in the form of RDF graphs were generated and uploaded to a semantic data store. 

An overview of the structure of the BBOB benchmark data format was presented in Section~\ref{sec:data_formats}. However, to properly annotate the performance data, we need to look at the specifics of the corresponfing file formats. First, the performance data is indexed by a function evaluation: for each evaluation that improves the objective function, a line gets written in the results file, containing the evaluation number, the raw objective value, and the transformed objective value (i.e., the value that is returned to the algorithm during the optimization process). 
The raw objective values are needed to allow for a fair comparison between different instances of the same function, while the transformed values are helpful when trying to reconstruct the input which was given to the optimizer. 

In addition to these two values, their best-so-far equivalents are also stored. When dealing with noiseless optimization and only writing data on function improvement, these values are redundant, but they can be helpful in other cases such as noisy optimization. 

Not all considered algorithms have data available on the same function/dimension/instance combination. This is partly caused by the shifting requirements of the BBOB workshops; i.e., the set of recommended instances and the number of repetitions per instance have not been identical throughout the years. In addition, some algorithms have been run only on a subset of the available BBOB collection, e.g., because of limited computational resources available. Since most of the algorithms benchmarked on BBOB are stochastic, there is a certain degree of variance between the runs.

The data in the ontology provides the terms/classes needed for annotation on the used problem/dimension/instance/algorithm in the corresponding benchmark analysis. In addition, data provenance information has been manually collected, linked to the performance data to trace its origin, and uploaded in the OPTION KB. The stored data provenance information includes the digital object identifier (DOI) of the paper where the experiments have been presented, the paper's title, the authors' name, and the year of publication. It is therefore possible to filter the data with respect to these criteria. 

\subsection{Example annotations of COCO-BBOB performance data}
\begin{figure*}[h!]
    \centering
    \includegraphics[width=0.85\textwidth]{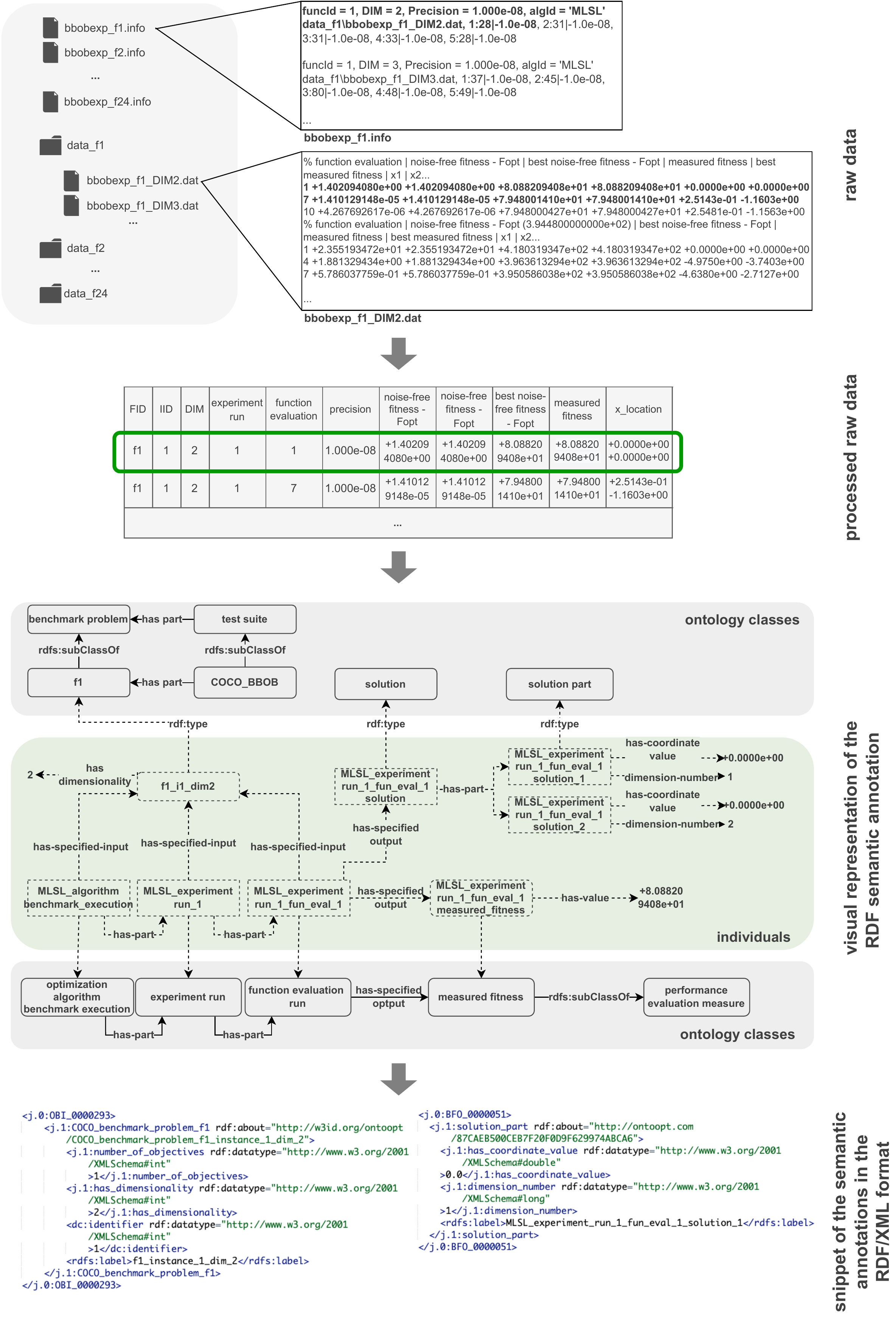}
    \caption{An illustrative example of semantic annotation of COCO-BBOB performance data.}
    \label{fig:exampleAnnotations}
\end{figure*}

In Figure~\ref{fig:exampleAnnotations}, we provide an example of a semantic annotation of performance data for the MLSL algorithm, which was benchmarked on the BBOB test suite using the COCO platform. We illustrate the process of creating OPTION-based semantic annotations in the RDF format.

As previously described, COCO separates performance data into multiple files and folders. More specifically, for each benchmark problem, there is a separate .info text file, which includes meta-information about the runs, and .dat files containing the raw performance data for these runs. 

We created a parser for COCO-formatted files and merged the information from the different files into a single table (see the processed raw data table in Figure~\ref{fig:exampleAnnotations}.

The processed performance data is then passed to the semantic annotation pipeline, that generates instances (also called individuals) of the OPTION classes. The semantic annotations are saved in the RDF format (which has a graph-like structure), where each fact is expressed in the form of subject - predicate - object triple. The subject and object are represented as nodes in the graph, with the predicate forming an edge between them. For instance, f1\_i1\_dim2 - rdf:type - f1 is an RDF triple denoting the fact that the problem instance labeled as f1\_i1\_dim2 is of type f1, where f1 is a class in the OPTION ontology representing the first benchmark problem from the BBOB test suite.

At the bottom of Figure~\ref{fig:exampleAnnotations}, a portion of the RDF annotations expressed in the RDF/XML syntax is displayed. Finally, the RDF/XML files are uploaded to a semantic data store (or a triple store) where they can be queried.

\subsection{Nevergrad}
The second use case covers semantic annotation of benchmark data obtained from Nevergrad - an open-source platform for black-box optimization \cite{nevergrad}. Nevergrad provides different test suites to benchmark the optimization algorithms. In the OPTION KB, we included annotations of 32 optimization algorithms benchmarked on the ten test suites YABBOB, YABIGBBOB, YACONSTRAINEDBBOB, YAHDBBOB, YAHDNOISYBBOB, YAHDSPLITBBOB, YANOISYBBOB, YAPARABBOB, YASMALLBBOB, and YASPLITBBOB. YABBOB~\cite{liu2020versatile} is a benchmark suite for black-box optimization problems inspired by the COCO-BBOB test suite. Moreover, in Nevergrad, there are different counterparts of YABBOB. For example, the YANOISYBBOB, YAHDBBOB, YAPARABBOB, YABIGBBOB variants of YABBOB contain problems with noise, high-dimensional, parallel, and big computational resources, respectively. Each of the test suites consists of 21 benchmark problems. The data is publicly available at \url{https://dl.fbaipublicfiles.com/nevergrad/allxpsnew/list.html}. 

There are several differences between the Nevergrad and COCO-BBOB benchmark data. First, the data obtained from the Nevergrad platform is stored at a coarser granularity level. Essentially, only the quality of the final solution is recorded, along with the experimental setup (budget of function evaluations, properties of the problem, etc.), and it is not recorded at the level of each function evaluation as is the case with COCO-BBOB. Also, no provenance data is available for the Nevergrad performance data. On the other hand, Nevergrad records offer other information that is lacking in COCO-BBOB (e.g., number of workers when running function evaluation in parallel and noise level of the function).

The differences in the performance data are inevitable as different benchmark platforms have different data formats. However, it is essential to note that the OPTION ontology was developed with special care not to be biased towards specific benchmark platforms. The annotation schema we propose is flexible enough to be used for semantic annotation of benchmark data from various platforms. Indeed, while there may be some information in other platforms that has not been considered while designing the ontology, the annotation schema can be easily extended to cover those aspects without affecting previously annotated data. 

\subsection{Landscape data}
In the third use case, we demonstrate the use of the OPTION ontology for the annotation of problem landscape data. For that purpose, we use a publicly available dataset~\cite{quentin_renau_2020_3886816} that contains the ELA features calculated for the first five instances of the 24 BBOB noiseless functions from the COCO environment in dimensions $D\in\{5, 10, 15, 20, 25, 30\}$. The 46 ELA features come from six feature groups (dispersion, $y$-distribution, meta-model, information content, nearest better clustering, and principal component analysis).
Since ELA features are not absolute and depend on the sampling strategy and the sample size~\cite{RenauDDD20PPSN}, this information is also added to the knowledge base. In our knowledge base, we include ELA features calculated with five different sampling strategies (i.e., LHS, iLHS, Random, Sobol, Randu) with $30D,50D,100D,250D,650D,800D,1000D$ sample sizes on a total of 100 independent repetitions. In addition, we store the median ELA feature value across these 100 repetitions.

Calculating the ELA features is a time-consuming task. Having calculated features in a format that automatically links them to the corresponding problems and enables easy access and querying is a large step towards more reusable research.

\section{The OPTION system for annotation, storage and querying}\label{sec:system}
In this section, we describe the OPTION ontology-based system we developed for semantic data management and integration.

The design of the system is governed by the goal and requirements presented in Section~\ref{sec:competency}. It currently supports the following 4 functionalities: (i) pipeline for semantic annotation of COCO-BBOB performance and landscape data and Nevergrad performance data; (ii) storage of the annotations in a RDF triplestore; (iii) REST API for querying the annotations and integrated query component in the IOHprofiler environment; (iv) web-interface for enabling users to contribute to OPTION and to the OPTION KB and to upload their own COCO-BBOB and Nevergrad data that will be semantically annotated. 

\subsection{The OPTION KB: annotation and storage}
\label{sec:technologies}
\begin{figure}[t]
    \centering
    \includegraphics[width=\linewidth]{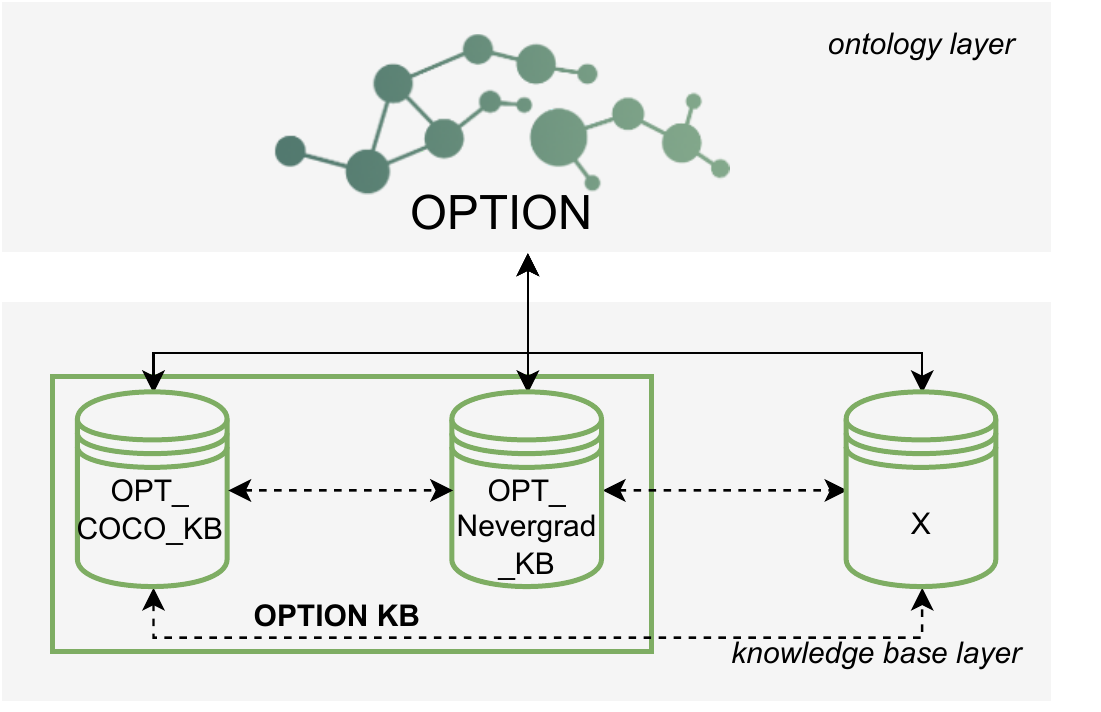}
    \caption{The OPTION ontology and the OPTION-aligned knowledge bases. Solid arrows signify the knowledge bases' explicit alignment with the ontology, which is accomplished through semantic annotation of the data. The interoperability of the various knowledge bases is denoted by dashed arrows, which is a direct result of the use of OPTION as a common vocabulary for annotation of the heterogeneous, distributed data.
 }
    \label{fig:KB}
\end{figure}

The OPTION ontology contains the semantic model, represented in a formal and standardized way. The OPTION KB, on the other hand, leverages the power of the ontology and holds the actual data that has been semantically annotated. In Section~\ref{sec:use-cases}, we discuss three use-cases of OPTION for integration of COCO-BBOB performance, Nevergrad performance, and COCO-BBOB landscape data. For that purpose, we have created two separate KB instances, OPT\_BBOB\_KB and OPT\_Nevergrad\_KB, that comprise the OPTION KB (see Figure~\ref{fig:KB}) and store the respective semantically annotated data. 
For semantic annotation of the raw COCO-BBOB and Nevergrad data, we developed scripts to parse the data, and created the semantic annotations using the Apache Jena RDF API.\footnote{Apache Jena RDF API: \url{https://jena.apache.org/documentation/rdf/index.html}} Once the annotation process is completed, the produced RDF annotations are uploaded to the Apache Jena TDB2 triple store. 

The BBOB and Nevergrad instances are on the same data server. However, that does not prevent other practitioners in the field of evolutionary computation from creating new KBs hosted on other servers. If they use the same vocabulary, the interoperability between the KBs is assured, meaning that they can be queried simultaneously if the data from multiple KBs is merged. 

\subsection{The OPTION KB: querying semantic annotations}
For querying the OPTION KB, we can use the SPARQL query language. We have set up an Apache Jena Fuseki2 server that connects to the Apache Jena TDB2 triple store and implemented two services to enable this functionality. The query service provides an endpoint for handling SPARQL queries in a RESTful manner~\cite{richardson2008restful}, while the upload service enables the upload of RDF data into the triple store. 

\begin{figure*}[b]
    \centering
    \includegraphics[width=\linewidth]{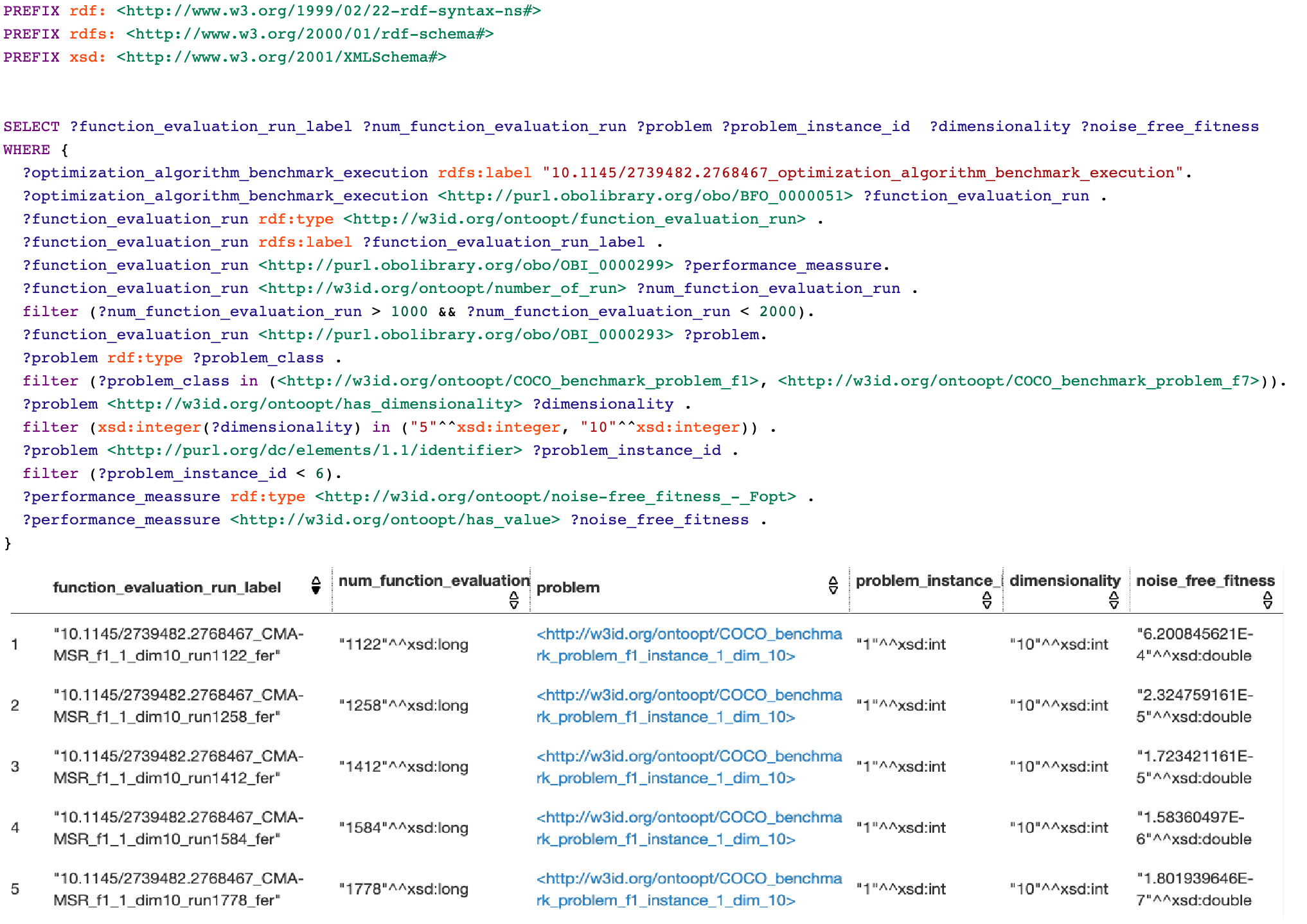}
    \caption{A screenshot from the FUSEKI query endpoint, presenting an example SPARQL query (at the top) and the first 5 answers to the query (at the bottom).}
    \label{fig:query}
\end{figure*}

In Figure~\ref{fig:query}, we present the listing of the SPARQL query for the following query expressed in natural language:
\begin{quote}
\textit{For all algorithms included in the study with DOI 10.1145/2739482.2768467 and for a fixed budget scenario with 1000-2000 function evaluations, return the \textit{noise-free fitness - Fopt} performance evaluation measure calculated on the first five instances of the f1 and f7 benchmark problems from the BBOB benchmark suite.}    
\end{quote} 
In addition, the bottom part of Figure~\ref{fig:query} shows the first five matches/answers for the same query. 

The query service supports all OPTION competency questions (and combinations of them), presented in Table~\ref{tab:competency}.

\subsection{Integration of the OPTION Knowledge Base in the IOHprofiler Environment}
As we can observe, SPARQL queries can become very complex and sometimes are seen as a bottleneck to the broader acceptance of Semantic Web technologies. We recognize that SPARQL query construction is an error-prone and time-consuming task that requires expert knowledge of the whole stack of semantic technologies. Even experts find it sometimes challenging to query semantic data since they first must get familiar with the data annotation schemes or the structure of the knowledge base.

\begin{figure}[t]
    \centering
    \includegraphics[width=0.5\textwidth]{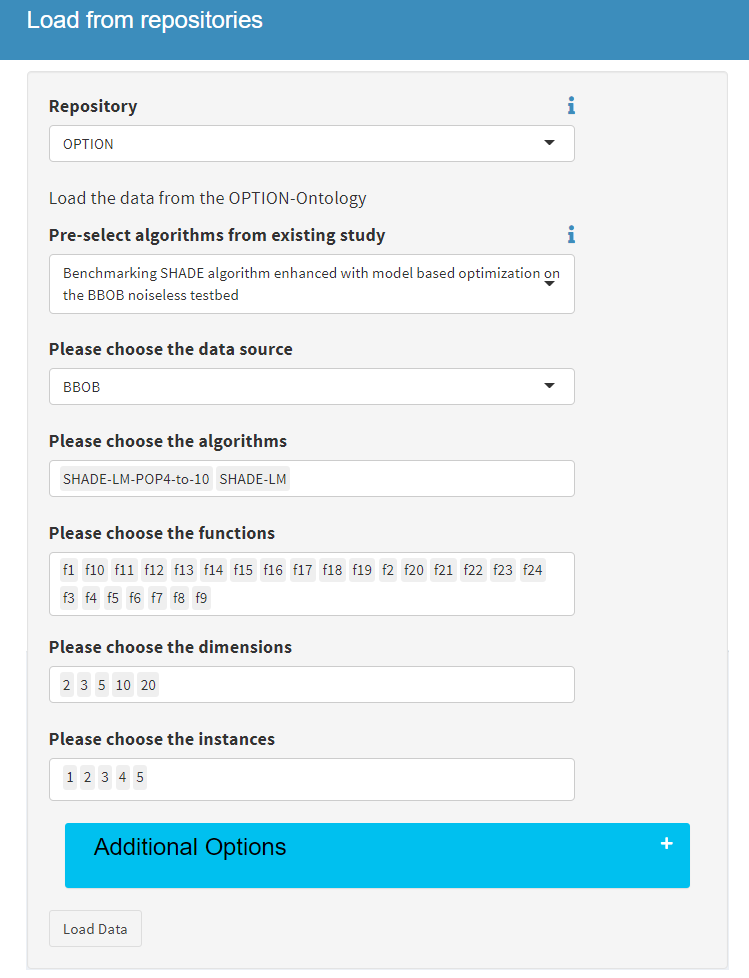}
    \caption{The interface of the OPTION-ontology queries within IOHanalyzer (version 1.6.3, available at \url{https://iohanalyzer.liacs.nl/}.)}
    \label{fig:IOH_interface}
\end{figure}

To facilitate the use of the OPTION ontology, we provide a simple GUI that can be used to gain access to performance data without needing to write SPARQL queries. This interface is connected directly to IOHanalyzer~\cite{IOHanalyzer}, which enables the loaded data to be used directly in performance analysis and visualization, and even be compared to data that might not yet be included in OPTION or to user-submitter performance data. Furthermore, the GUI provides access to a parameterized search process, which can be used without any underlying knowledge about the used semantic data model. Users can express their query by selecting from several drop-down options, which specify the required information, such as suite, function, algorithm, etc., and load the corresponding performance data to analyze. This interface is shown in Figure~\ref{fig:IOH_interface}. While this interface is static, it illustrates the power of integrating the ontology into IOHanalyzer: users without any background knowledge can use it to gain insight into the performance of the selected algorithms/functions. 

Additionally, this interface can be easily expanded based on the community's wishes. To illustrate this potential, we created another entry point into OPTION, which can be used to load all performance data which originated in a specified paper. To this end, the user selects a paper by its title, which then populates the relevant information about the used algorithms and functions in that study. 
By loading this pre-selected data, the user has full access to the performance data of the selected study, which they can then investigate in more detail by making use of the visualizations within IOHanalyzer. This type of interactive analysis then allows the user to look at the data from different perspectives and to compare it to other algorithms. 

\subsection{Extending the OPTION ontology and knowledge base}
So far, the OPTION ontology has been successfully applied for data integration and management tasks from the COCO and Nevergrad benchmark platforms. Data that was previously stored in different data formats and that could not be queried, is now integrated and can be queried simultaneously. 
However, extending the ontology is not a trivial task, as it requires its contributors to have a good understanding of the semantic model. Also, the process of annotating performance data from an arbitrary platform, hence populating the knowledge base, currently cannot be fully automated. 

\begin{figure*}[h]
    \centering
    \includegraphics[width=0.8\linewidth]{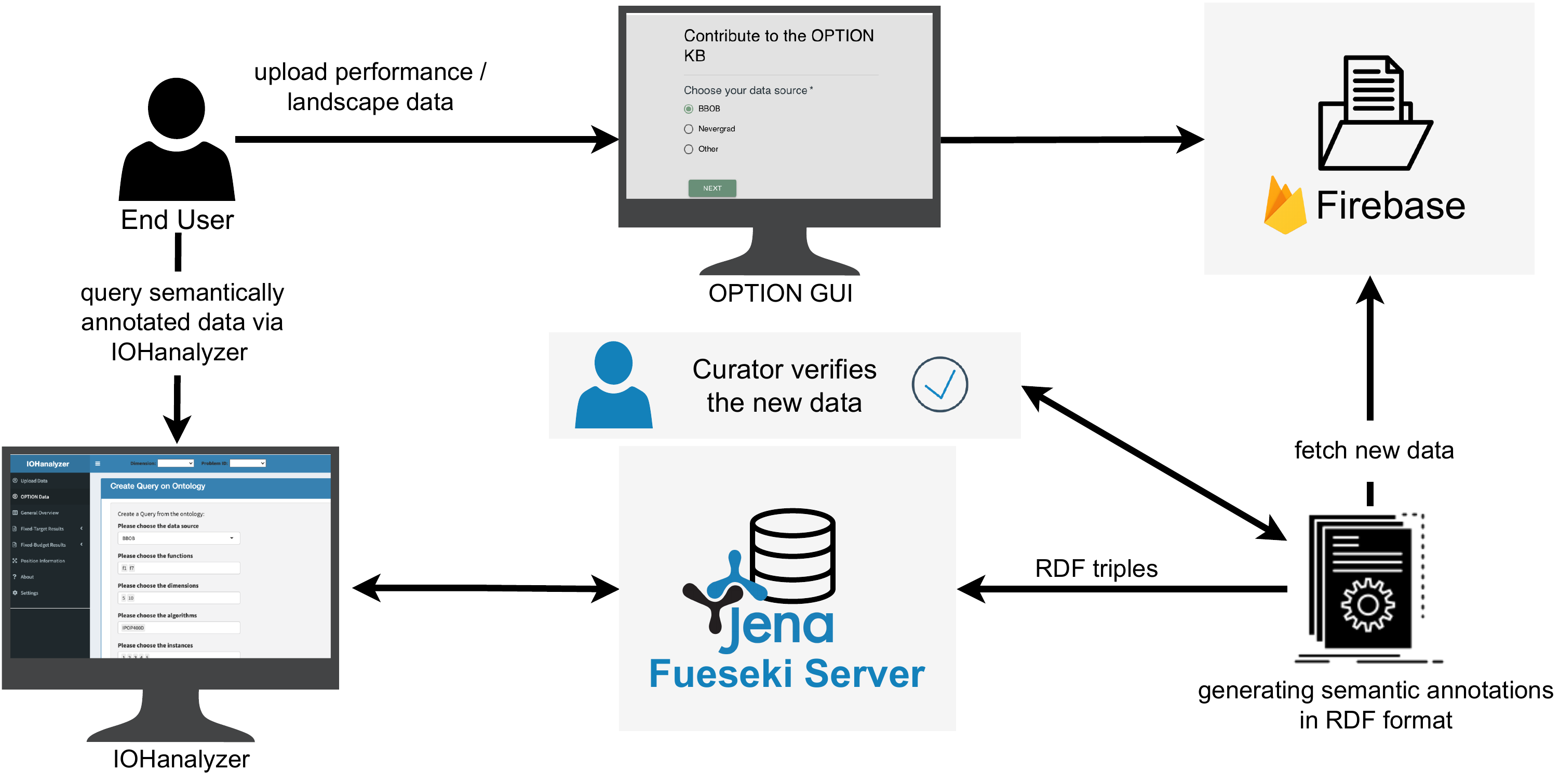}
    \caption{A flowchart of the process of uploading, annotating, and querying new data in the OPTION KB. }
    \label{fig:flowchart}
\end{figure*}

To facilitate the uploading of new data to the OPTION KB, we have developed a web interface available at: \url{http://semantichub.ijs.si/OPTION/}. Currently, the web interface supports the uploading of COCO-BBOB and Nevergrad performance and landscape data from published studies. Figure~\ref{fig:flowchart} depicts the process of submitting new data, semantically annotating it, integrating the annotations with the OPTION KB, and querying it. Via the web interface, end users can first upload the raw data, details about the study, and related provenance information. 

The uploaded data is stored on a Firebase server. Then, periodically,
we retrieve the newly uploaded data and semantically annotate it
In order to ensure the high quality of the OPTION KB, we include a curator in the loop who prior to executing the semantic annotation pipeline, verifies that the data format conforms to the COCO/Nevergrad data format and verifies the study-related provenance metadata provided by the user.

Finally, requests for extending the OPTION ontology and knowledge base can be made directly via the web interface or via the GitHub repository\footnote{OPTION at GIT: \url{https://github.com/KostovskaAna/OPTION-Ontology}}, after which we can establish a collaboration and help guide the whole extension process. We encourage researchers and especially developers of benchmark platforms to adopt the use of the OPTION ontology, to annotate their benchmark and problem landscape data based on the OPTION ontology, and maintain their own OPTION-aligned knowledge bases, as depicted in Figure~\ref{fig:KB}. Distributed knowledge bases based on same ontological vocabulary can be easily queried using federated querying strategies\footnote{SPARQL 1.1 Federated Query: \url{https://www.w3.org/TR/sparql11-federated-query/}}.

\section{Conclusions \& Future work}\label{sec:conclusions}
The current version of the OPTION ontology has been developed from a performance and problem-centered perspective. This perspective allows it to effectively handle the most common types of queries that arise in the analysis of benchmark data. However, this also means that information about the specific algorithms is somewhat limited. The lack of information about algorithms is partly due to the inaccessibility of this type of meta-information: common benchmarking setups only store high-level features about algorithm settings. To further extend the ontology presented in this paper, we aim to expand the knowledge base for these algorithm-specific details. Moreover, a more detailed semantic representation of the algorithm space should be included to describe the algorithm family, operators, hyperparameters, etc. Recently, several studies have attempted to unify taxonomies over the algorithm space~\cite{stegherr2020classifying, stork2020new, liu2021survey}, but further work is needed to develop a general structure that can be incorporated into OPTION.

The development of the OPTION ontology is a step forward in improving the reusability and interoperability of performance and problem landscape data. By annotating a large subset of BBOB, Nevergrad, and ELA data, we have demonstrated the potential of the ontology to support data integration while providing powerful query capabilities for direct analysis of the required datasets. This significantly reduces the time required to collect data across many functions and algorithms, while providing flexibility in managing the performance perspective (i.e., fixed budget, fixed target).

Since the development of the OPTION ontology is motivated by the need for a single model that connects different data formats to make them fully interoperable, we plan to annotate benchmark data coming from several actively used benchmarking platforms. Our future work will also include the modular representation of algorithms and the use of ontology reasoners to infer the taxonomies based on the Description Logics rules we will provide. This would allow new algorithms or problems to be added by simply answering a set of predefined questions, allowing the ontology to grow organically. The creation of reproducible and readily available data will ultimately benefit the entire optimization community, so the effort required to achieve this goal would be very worthwhile.

While we plan to keep the ontology-based system for annotation, storage and querying up-to-date to be relevant for the evolutionary computation research community, we encourage researchers and especially developers of benchmark platforms to adopt the use of the OPTION ontology, annotate their own benchmark and problem landscape data, and maintain their own knowledge bases.

Finally, we plan to employ OPTION in a predictive study based on machine learning. The OPTION KG will be used to create a knowledge graph (KG) comprising two types of nodes, problem instances, and optimization algorithms. An edge/link between two such nodes would mean that the optimization algorithm can reach a given target precision for a given budget. An ML task to be considered would be to predict links between new optimization algorithms and problem instances by employing graph-based and KG embedding methods, together with ML methods. This would offer a novel approach to understanding the performance of optimization methods.

\section*{Acknowledgment}
We acknowledge the support of the Slovenian Research Agency through program grants No. P2-0103 and P2-0098, project grants No. J2-9230 and N2-0239, and a young researcher grant to AK, as well as the EC through grant No. 952215 (TAILOR). Our work is also supported by Paris Ile-de-France region, via the DIM RFSI AlgoSelect project, and via a \href{http://species-society.org/scholarships-2022/}{SPECIES scholarship} for Ana Kostovska.

\ifCLASSOPTIONcaptionsoff
  \newpage
\fi

\bibliographystyle{IEEEtran} 
\bibliography{references}

\end{document}